\documentclass{article}

\usepackage{PRIMEarxiv}

\usepackage[utf8]{inputenc} 
\usepackage[T1]{fontenc}    
\usepackage{hyperref}       
\usepackage{url}            
\usepackage{booktabs}       
\usepackage{amsfonts}       
\usepackage{nicefrac}       
\usepackage{microtype}      
\usepackage{lipsum}
\usepackage{fancyhdr}       
\usepackage{graphicx}       
\graphicspath{{media/}}     

\title{Cattle Identification Using Muzzle Images and Deep Learning
Techniques
}

\author{
  Geoffrey Kimani \\
  CyLab-Africa / Upanzi Network \\
  Kigali\\
  \texttt{gnkimani@andrew.cmu.edu} \\
   \And
  Oluwadara Adedeji \\
  CyLab-Africa / Upanzi Network \\
  Kigali\\
  \texttt{oadedeji@andrew.cmu.edu} \\
  \And
  Patrick Fashingabo \\
  CyLab-Africa / Upanzi Network \\
  Kigali\\
  \texttt{pfashin2@andrew.cmu.edu} \\
  \AND
   Moise Busogi \\
   Carnegie Mellon University Africa  \\
 \texttt{mbusogi@andrew.cmu.edu} \\
  \And
  Edith Luhanga \\
  Carnegie Mellon University Africa  \\
  \texttt{eluhanga@andrew.cmu.edu} \\
  \And
  Karen Sowon \\
  Carnegie Mellon University  \\
  \texttt{scheruto@andrew.cmu.edu} \\
  \And
  Lenah Chacha \\
  CyLab-Africa /\ Upanzi Network  \\
  \texttt{lchacha@andrew.cmu.edu} \\
}

\begin{document}
\maketitle

\begin{abstract}
Traditional animal identification methods such as ear-tagging, ear notching, and branding have been effective but pose risks to the animal and have scalability issues. Electrical methods offer better tracking and monitoring but require specialized equipment and are susceptible to attacks. Biometric identification using time-immutable dermatoglyphic features such as muzzle prints and iris patterns is a promising solution. This project explores cattle identification using 4923 muzzle images collected from 268 beef cattle. Two deep learning classification models are implemented - wide ResNet50 and VGG16\_BN and image compression is done to lower the image quality and adapt the models to work for the African context. From the experiments run, a maximum accuracy of 99.5\% is achieved while using the wide ResNet50 model with a compression retaining 25\% of the original image. From the study, it is noted that the time required by the models to train and converge as well as recognition time are dependent on the machine used to run the model.

Code files
\footnote{Code files: {\url{https://github.com/peter716/Animal_Biometrics_System}}}
and data files.
\footnote{Data files: {\url{https://zenodo.org/record/6324361\#.ZAsiTNJBzb3}}}

\end{abstract}

\section{Introduction}
\subsection{Background}
In the realm of animal identification, classical and electrical methods have long been employed, but they possess certain shortcomings. Classical methods involve permanent or temporal identification techniques. Permanent methods, such as ear notching, freeze branding, and hot-iron branding, leave permanent marks on animals but are often invasive and cause pain. Temporal methods, like ear-tagging, are less painful but susceptible to fraud and manipulation. These methods require manual work, animal restriction, and can be unreliable in preventing fraud or duplication \cite{ref1, ref2, ref3, ref4}
.

Electrical methods, such as ear tags, injectable transponders, and rumen boluses, rely on electronic devices to identify and track cattle. While these methods are less invasive and easy to apply, they can cause discomfort to the animals and require additional equipment, maintenance, and calibration. Moreover, they can be susceptible to fraud and attacks \cite{ref1, ref2, ref4}.

To overcome the limitations of classical and electrical methods, biometric techniques have gained prominence in animal identification. Biometric modalities utilize unique physiological or behavioral features to assign a distinct identity to each animal \cite{ref1}. For bovine animals, these features can include retinal vascular patterns, iris patterns, or muzzle prints.

Retinal vascular patterns are based on the unique blood vessel patterns in the retina. They provide accurate identification, are relatively unaffected by corneal injuries, and can be used in various animals. However, certain inherited retinal diseases and congenital ocular disorders may impact the recognition process. Acquiring high-quality retinal images using techniques like optical coherence tomography (OCT) allows the extraction of features such as vessel diameter, branching patterns, and curvature. These features are then used to create a unique template for matching and identification \cite{ref1,ref12, ref13}.

Iris patterns, found in the colored part of the eye, are another biometric feature. They are determined by the distribution of pigments in the iris, resulting in a stable and distinctive identifier. However, iris patterns can be affected by diseases and aging, which may reduce accuracy. Extracting iris texture, color, and intensity using techniques like Gabor filters enables the creation of unique templates for recognition \cite{ref1, ref11, ref14, ref15}.

Muzzle prints, analogous to human fingerprints, offer an accurate and time-immutable identifier. High-quality images of the muzzle can be used to extract unique patterns of beads and ridges. While the effect of diseases on muzzle prints is not extensively studied, conditions altering the textual structure of the muzzle may affect identification. Templates generated from muzzle prints are compared to verify an animal's identity \cite{ref1}.

Various approaches have been proposed to implement biometric identification systems. Some employ deep learning techniques, such as convolutional neural networks (CNNs) or deep belief networks (DBNs), to process muzzle images. These models achieve high accuracies, but they often require large amounts of data and computational resources. Other approaches utilize template matching by creating databases of unique templates extracted from biometric features. These templates are compared with incoming data to determine a match or similarity score, enabling identification \cite{ref4, ref5, ref6, ref7, ref8}.

Biometric methods offer non-invasive, reliable, and accurate means of identifying animals. Techniques such as retinal vascular patterns, iris patterns, and muzzle prints provide distinct features for identification. Deep learning models and template matching approaches have shown promising results in animal identification systems, offering improved accuracy and non-invasiveness compared to classical and electrical methods.

\subsection{Problem statement}
Animal identification has been a problem long addressed in various ways with the aim of accurately recognizing individual animals. The identification techniques have been broadly categorized into four classes: permanent methods, temporary methods, electrical methods, and animal biometrics \cite{ref1}. In the present study, permanent and temporary methods are amalgamated into a single category termed \textit{classical methods}. This category encompasses practices such as ear-tagging, ear notching, ear tattooing, hot iron branding, and freeze branding. However, these methods have serious shortcomings, including fraud, infections, animal distress, equipment loss (ear tags), and limited scalability \cite{ref1, ref2}. Electrical methods, characterized by rumen boluses, Radio Frequency Identification (RFID) ear tags, and injectable glass tags \cite{ref1}, offer advantages over classical methods by addressing tracking and animal monitoring beyond identification requirements. However, they come with high installation costs due to specialized equipment and personnel. Moreover, electrical methods are vulnerable to Denial of Service (DoS) attacks, tag modification, and system spoofing \cite{ref1, ref2, ref4}. Biometric identification, leveraging time-immutable dermatoglyphic features, offers harmless animal identification. These methods provide automated authentication, high security, reliability, and zero memory load on humans \cite{ref1}. Muzzle prints, retinal vascular patterns, and iris patterns are among the various techniques used for biometric identification.

In the current study, biometric systems are examined for the purpose of cattle identification, specifically focusing on the unique identification of cattle through muzzle images. Previous research \cite{ref1} has explored three biometric identifiers for cattle: muzzle images, iris patterns, and retinal vascular patterns. The dataset employed in this study is identical to that used by \cite{ref8}, comprising 4923 muzzle images from 268 beef cattle \footnote{Beef Cattle Muzzle/Noseprint database for individual identification \label{datasource} \url{https://zenodo.org/record/6324361\#.ZAsiTNJBzb3}}. The study further narrows its focus to applications in resource-constrained environments where high-quality digital cameras may not be readily accessible.

\subsection{Use cases discussion}
Animal identification is very important in establishing ownership, preventing fraud, and identifying theft. In this section, various use cases are presented that could benefit from the implementation of a biometric identification system. These use cases are particularly challenging to address using traditional classical and electrical methods of identification. There are various applications where such a system could be useful including disease tracking, insurance claims, and collateral for financial loans among others. With the biometric systems being able to accurately identify individual cattle, banks, and insurance companies could find this of interest to them since they would be able to establish ownership and proof of claims for insurance or collateral for bank loans. This would suppress insurance fraud and allow farmers to use their cattle as security for bank loans. Government and research institutions dealing with animal production and animal health might find the system useful as well. A fully connected system with a front-end dashboard for user interaction and a database can be built on top of the proposed deep-learning model to keep track of vaccination programs as well as track animal disease trajectories. This information might also be useful to public health officials who need to certify meat that is sold for consumption. Security agencies might also find this system useful in cases of identifying missing or swapped cattle and establishing ownership in cases of dispute. It should be noted that the biometric system alone may not adequately address all challenges inherent to a specific use case. As such, it may require integration with other systems to achieve full utility. For example, in disease tracking, the system would ideally rely on an existing system that keeps track of ownership, region, vaccinations administered, animal movements and interactions(electrical methods are useful in this case), etc. whether it is an independent system or extension of the biometric system.

\section{Methods} \label{methods}
\subsection{Dataset}
\subsubsection{Data collection}
The dataset employed in the current study was sourced from previous research \cite{ref8}, comprising 4923 muzzle images from 268 beef cattle. The dataset is available at \ref{datasource} under Creative Commons license 4.0 and is provided as a directory of cattle images for each of the 268 cattle. Each cattle had at least 4 images representing it while some cattle had up to 70 images. These images were obtained from cattle in their natural environment while grazing under natural lighting conditions and a high-quality mirrorless digital camera with a 70–300 mm F4-5.6 focal lens. The muzzle areas in the images were first horizontally aligned and then cropped out to focus the muzzle area resulting in images uniformly resized to 300 by 300 pixels. It is important to note that though a cursory inspection of the muzzle prints may show that the color and texture of the prints look similar in different animals the patterns of beads, grooves, and ridges are different upon a comprehensive analysis.

\subsubsection{Data splitting}
For data preparation, the dataset was partitioned into 70\% for training and 30\% for evaluation. To guarantee the representation of each individual cattle in both the training and testing sets, an iteration was performed over the image folders, selecting a train and test set for each individual animal. This approach ensures that even the least represented animals are included in both the training and testing sets.

\subsubsection{Data augmentation}
In this section, various data augmentation techniques were applied, including image flipping, color jittering, blurring, and rotations. Images were resized to dimensions of 300 by 300 pixels. Random horizontal flips were introduced to augment the dataset by generating additional training examples with varied orientations, thereby enhancing the model's generalization performance. Color jittering was employed to adjust the brightness, contrast, saturation, and hue of the images, further contributing to the diversity of the training examples. Gaussian blur was applied to smooth out noise and pixelation, aiding in feature extraction and classification. Data augmentation was conducted using PyTorch modules.

\subsection{Modeling}
In the present study, two models were selected from the 59 deep-learning models proposed in the baseline research. The VGG16\_BN model was chosen due to its superior performance in the baseline paper, and wide ResNet50 was selected as an alternative, representing a newer architecture. A transfer learning approach was employed, utilizing models pre-trained on ImageNet. These models are readily available through PyTorch\footnote{\url{https://pytorch.org/vision/stable/models.html}}.

\subsubsection{Architecture Discussion}

The VGG16 model is a convolutional neural network architecture that was developed by the Visual Geometry Group at the University of Oxford \cite{ref16}. The model has 16 layers, consisting of 13 convolutional layers and 3 fully connected layers, and is known for its simplicity and effectiveness in image classification tasks. The VGG16 model architecture is characterized by a series of convolutional layers, and each convolutional layer applies multiple 3x3 convolutional kernels to the input feature maps to produce a set of output feature maps. The number of kernels in each convolutional layer increases as the network goes deeper, which allows the model to learn increasingly complex and abstract features. The max-pooling layers are used to reduce the spatial dimension of the feature maps. The last few layers of the network are fully connected layers that map the extracted features to a final set of output classes. In PyTorch, the VGG16 model can be imported and used as a pre-trained model for transfer learning, as described in the documentation \footnote{ \url{https://pytorch.org/vision/stable/models/generated/torchvision.models.vgg16.html}}. The pre-trained VGG16 model in PyTorch has been trained on the large-scale ImageNet dataset, which contains over 1 million images and 1000 classes \cite{ref17}. This pre-training allows the model to learn a set of general features that can be transferred to other image classification tasks with relatively little additional training. This is particularly useful when working with smaller datasets, as transfer learning can help to improve the performance of the model and reduce overfitting.

\begin{figure}
    \centering
    \begin{minipage}{0.5\textwidth}
        \includegraphics[width=\textwidth]{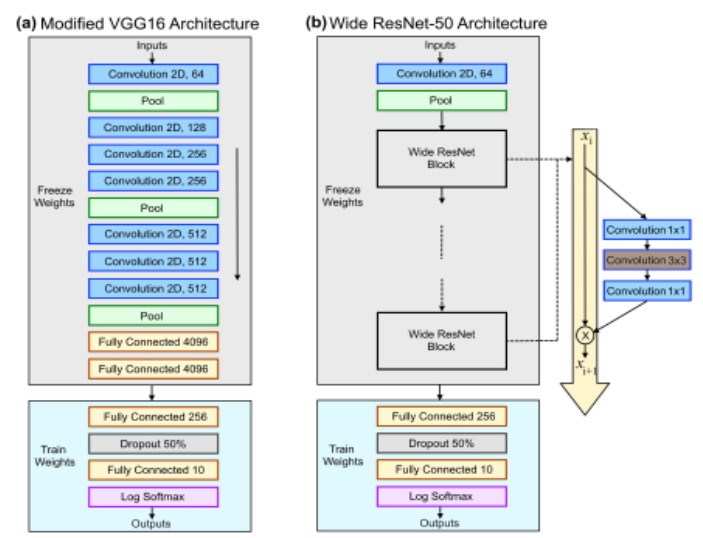}
        \caption{Pretrained Models Architecture \\ {\tiny (Source: Deep Transfer Learning for Land Use and Land Cover Classification: A Comparative Study, Raoof Naushad et. al.)}}
        \label{fig:model-arch}
    \end{minipage}
\end{figure}

The Wide ResNet architecture is based on the ResNet architecture, which is a deep convolutional neural network that uses residual connections to improve the performance of the network. It is similar to the ResNet architecture but uses wider layers with more filters to improve the accuracy of the network. The model starts with an input layer followed by a convolution layer with 64 filters and a kernel size of 7x7 which are then followed by batch normalization and ReLU activation layers finally a max pooling layer. This section is then followed by at least four residual blocks each containing  multiple convolution layers with 128, 256 or 512 filters. A global average pooling layer follows to reduce the spatial dimensions of the feature maps and finally followed by a fully connected layer and a softmax activation layer \cite{ref18}. In PyTorch, the pre-trained Wide ResNet model is available as part of the torchvision library
\footnote{%
  \begin{minipage}{\columnwidth}
\url{https://pytorch.org/vision/stable/models/generated/torchvision.models.wide_resnet50_2.html\#torchvision.models.wide_resnet50_2}
  \end{minipage}%
}
which provides a number of pre-trained models for use in computer vision tasks. Figure \ref{fig:model-arch} shows the architectures side by side 

\begin{figure}
    \centering
    \begin{minipage}{0.5\textwidth}
        \centering
        \includegraphics[width=\textwidth, height=10cm]{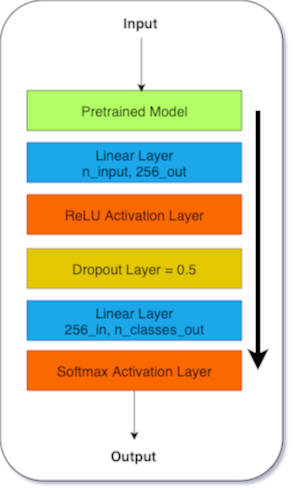}
        \caption{Modified Architecture}
        \label{fig:model-arch-our}
    \end{minipage}
\end{figure}

In the current study, modifications were made to the architecture to include a custom fully connected layer tailored to the specific number of classes, in this case, 268. A linear layer was added to the pre-trained model, accepting \( n \) inputs—where \( n \) is the number of classes output by the model, 1000 for \textit{IMAGENET1K\_V1}—and outputting 256 classes. This was followed by a ReLU activation layer to aid the model in learning the decision boundaries for different classes, as well as a dropout layer to mitigate overfitting. Subsequently, another linear layer was introduced, taking the 256 classes and outputting the 268 classes specific to this study. The architecture concludes with a softmax activation layer, providing a probability distribution over the predicted classes. Figure \ref{fig:model-arch-our} shows our new architecture skeleton.

\subsubsection{Image compression}

A primary objective of this study was to enhance model robustness in the context of low-quality images. This consideration is particularly relevant for potential applications in Africa, where acquiring high-quality images akin to those used in the baseline paper may be challenging. Furthermore, in real-world implementations, lower-grade cameras may be employed as opposed to high-quality ones, thereby affecting image quality. Consequently, the study sought to explore techniques that would facilitate experimentation with low-quality images, given the unavailability of a secondary dataset for this purpose.

\begin{figure}
    \centering
    \begin{minipage}{0.5\textwidth}
        \centering
        \includegraphics[width=\textwidth, height=8cm]{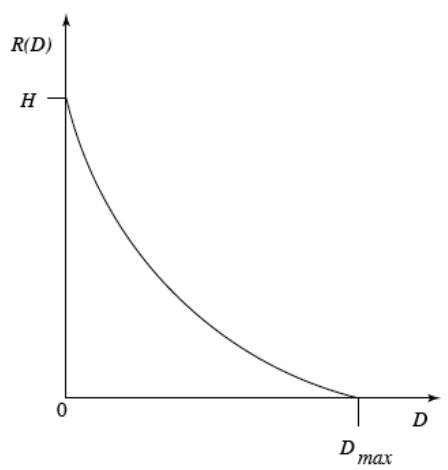}
        \caption{Compression scheme theory \\
        {\tiny (Source: BITMOVIN \url{https://bitmovin.com/lossy-compression-algorithms/})}
        }
        \label{fig:compression}
    \end{minipage}
\end{figure}

The study explored the use of image compression techniques. Compression in the context of information theory is the process of reducing the size of a data file by removing redundant or unnecessary information. This is an important process as it reduces the amount of information being transmitted. There are two main schemes in compression; lossy and lossless compression. Each scheme is determined by the data rate against the distortion level as shown in figure \ref{fig:compression}. The data rate refers to the amount of data needed to represent the compressed image usually measured in bits per second (bps) while distortion refers to the difference between the original and resulting data. On the x-axis is the distortion level and the y-axis is the data rate. If you have a high data rate and zero distortion, it is a lossless compression scheme, conversely, if you have a lower data rate in the resulting data than in the original and a distortion greater than 0 you have a lossy compression scheme. Our interest was the lossy compression scheme that allows us to lose some image information consequently losing some quality in the images to be compressed.

Lossy image compression has two steps - quantization and transform coding. The quantization step involves dividing the image into small, fixed-size blocks of pixels, and then mapping the color values of each block to a limited range of possible values. The mapping process discards some of the original color information, resulting in a loss of detail, but allows for smaller file sizes \cite{ref18}. In the coding step, the quantized data is compressed using algorithms that reduce the amount of information needed to represent the image. A common technique is to use transform coding, in which a mathematical transform is applied to the image data to reduce redundancy before encoding it \cite{ref18}. The information lost in this step is permanently lost and cannot be recovered. A good example of a lossy algorithm popular in image compression is the discrete cosine transform(DCT) algorithm. 

The DCT algorithm transforms the image data from the spatial domain (pixel values) to the frequency domain (frequency components that make up the image). In the frequency domain, the DCT separates the image data into a set of coefficients that describe how much different frequency components contribute to the image. The DCT algorithm is very efficient and is used in JPEG(Joint Photographic Experts Group) compression, a popular industry-grade image compression scheme.

\begin{figure}
    \centering
    \begin{minipage}{0.5\textwidth}
        \centering
        \includegraphics[width=\textwidth, height=5cm]{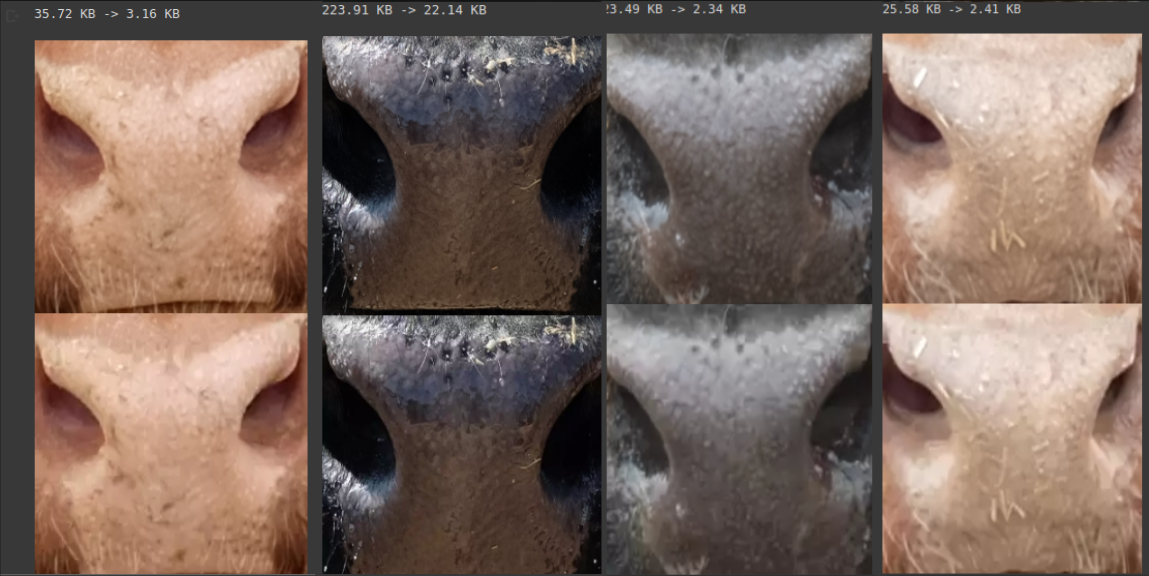}
        \caption{Original and Compressed Images Sample}
        \label{fig:compression-mash}
    \end{minipage}
\end{figure}

In addressing the issue, the study employed image compression as a strategy. While a custom implementation of the Discrete Cosine Transform (DCT) algorithm was initially considered, it proved to be time-inefficient. As a result, the study opted for the use of existing, optimized libraries. Specifically, the Pillow library was utilized for its image compression capabilities. This library allows for the specification of image quality during the save operation. For instance, specifying a quality of 50 results in an image retaining 50\% of its original quality. The library also offers an optimize parameter for further file size reduction. Pillow is a fork of the Python Image Library (PIL). In this study, images were compressed to retain three different quality levels—100\%, 50\%, and 25\%—and the model's performance was evaluated at each of these quality levels. Figure \ref{fig:compression-mash} shows sample images compressed at 50\% of the original quality. The images on the top row are the original while the bottom are the compressed ones. A visual cursory examination of the images may not show huge differences but the crispness, sharpness, and color differentiation of object edges within the image are reduced (possibly not apparent to the eye) by compression.

\section{Results}

A series of experiments were conducted to assess the performance of the selected models. It was observed that the wide ResNet50 model achieved the highest accuracy, approximately 99.4\%, outperforming the baseline's highest accuracy of 98.7\% achieved with the VGG16\_BN model. Early stopping mechanisms were implemented to facilitate quicker model convergence. Notably, both selected models converged in fewer than 40 epochs, which is more efficient than the 50 epochs reported in the baseline study.  

It is noted that the runtime of the models is dependent on the machine spec as seen in tables \ref{tab:resnet50} and \ref{tab:vgg16}. These experiments were run on two different machines both provided under the Google Colab platform. Results in \ref{tab:resnet50} were obtained using the free account which comes with an Intel Xeon CPU @2.20 GHz, 13 GB RAM, Tesla K80 accelerator, and 12 GB GDDR5 VRAM. Results in \ref{tab:vgg16} were obtained using a paid version of colab featuring more powerful specs Intel(R) Xeon(R) CPU @ 2.30GHz Tesla P100-PCIE-16GB accelerator and 12 GB GDDR5 VRAM with a high-RAM option available. The difference in these machines explains the difference in the total time taken as well as time per epoch between the two experiments. With more powerful machines the time per epoch and total time decreases considerably. For example in a separate experiment using an NVIDIA A100-SXM4-40GB machine, it took 25 seconds per epoch with a total of 30 epochs run bringing the total time to around 12.5 minutes. 

The performance of the models actually improved after the lossy compression but it was noted that the epochs required to converge are more. Upon examining Table \ref{tab:resnet50}, it is observed that when retaining 100\% of the original image quality, the model converges after 28 epochs, achieving an accuracy rate of 99.2\%. When 50\% of the image quality is retained, it takes 8 more epochs to converge but a higher accuracy of 99.4\% is attained while retaining 25\% of the original quality the model converges after 30 epochs with an accuracy of 99.5\%. The same is true for the VGG16\_BN model as a similar trend is observed. Given that the images are of lower quality, the models take longer to learn the features and this could explain the additional epochs. 

In terms of accuracy measurement, both the initial accuracy achieved after the first epoch and the final accuracy upon model convergence are reported. These values are presented in the accuracy column of the respective tables, with the top value indicating the starting accuracy and the bottom value representing the final accuracy. The VGG16\_BN model starts with a very low accuracy but also attains a lower accuracy than the wide ResNet50 model. While quality is reduced, the images are quite large so the models are still able to extract useful features. 

The time taken by each of the models was also documented.Experiments were conducted on two different platforms: Colab and Colab Pro. While the exact machine specifications are not available due to Colab's dynamic assignment, it is noted that Colab Pro generally offers more powerful machines. Wide ResNet50 required a longer total training time compared to VGG16\_BN, which was run on Colab Pro. Interestingly, the total time taken was nearly similar across different levels of image compression. However, higher compression levels did result in longer training times. Regarding time per epoch, compressed images led to shorter epoch times for both models. It was also observed that the more powerful machine allocated for VGG16\_BN on Colab Pro resulted in significantly shorter epoch times compared to those for Wide ResNet50.

Tables \ref{tab:resnet50} and \ref{tab:vgg16} summarize the results from the two models comparing the performance based on image quality retained.

\begin{table*}[!t]
\scriptsize
\centering
\begin{tabular}{|l|p{1.2cm}|p{1.3cm}|p{1.5cm}|p{1.5cm}|p{1.5cm}|p{1.5cm}|}
\hline
\textbf{\centering Experiment} &
\textbf{\centering File sizes} &
\textbf{\centering Percent of the Original Quality} &
\textbf{\centering Accuracy} &
\textbf{\centering Epochs \\ (Early stopping)} &
\textbf{\centering Total Time} &
\textbf{\centering Time per epoch (mins)} 
\\ \hline
\textbf{Normal run} &
622MB (4923) &
100\% &
{ \begin{tabular}[c]{@{}c@{}}9\%\\ $\sim$99.2\%\end{tabular}} &
28 &
\begin{tabular}[c]{@{}c@{}}1 hr \\ 12 min\end{tabular} &
$\sim$3.2
\\ \hline
\textbf{Pill JPEG Compression} &
\begin{tabular}[c]{@{}c@{}}76MB\\ (4923)\end{tabular} &
50\% &
{ \begin{tabular}[c]{@{}c@{}}7\%\\ $\sim$99.4\%\end{tabular}} &
36 &
1 hr 29 min &
$\sim$2.4
\\ \hline
\textbf{Pill JPEG Compression} &
\begin{tabular}[c]{@{}c@{}}45MB\\ (4923)\end{tabular} &
25\% &
{ \begin{tabular}[c]{@{}c@{}}7\%\\ $\sim$99.5\%\end{tabular}} &
30 &
\begin{tabular}[c]{@{}c@{}}1hr\\ 10 min\end{tabular} &
$\sim$2.3
\\ \hline
\end{tabular}
\caption{Wide ResNet50 Results}
\label{tab:resnet50}
\end{table*}

\begin{table*}[!t]
\scriptsize
\centering
\begin{tabular}{|l|p{1.2cm}|p{1.3cm}|p{1.5cm}|p{1.5cm}|p{1.5cm}|p{1.5cm}|}
\hline 
\textbf{\centering Experiment} &
\textbf{\centering File sizes} &
\textbf{\centering Percent of the Original Quality} &
\textbf{\centering Accuracy} &
\textbf{\centering Epochs \\ (Early stopping)} &
\textbf{\centering Total Time} &
\textbf{\centering Time per epoch (mins)} 
\\ \hline
\textbf{Normal run} &
622MB (4923) &
100\% &
{ \begin{tabular}[c]{@{}c@{}}0.7\%\\ $\sim$97.7\%\end{tabular}} &
38 &
\begin{tabular}[c]{@{}c@{}}53 min\end{tabular} &
$\sim$1.4 \\ \hline
\textbf{Pill JPEG Compression} &
\begin{tabular}[c]{@{}c@{}}76MB\\ (4923)\end{tabular} &
50\% &
{ \begin{tabular}[c]{@{}c@{}}0.6\%\\ $\sim$97.97\%\end{tabular}} &
39 &
53 min &
$\sim$1.38 \\ \hline
\textbf{Pill JPEG Compression} &
\begin{tabular}[c]{@{}c@{}}45MB\\ (4923)\end{tabular} &
25\% &
{ \begin{tabular}[c]{@{}c@{}}0.4\%\\ $\sim$98.04\%\end{tabular}} &
43 &
\begin{tabular}[c]{@{}c@{}}58 min\end{tabular} &
$\sim$1.35 \\ \hline
\end{tabular}
\caption{VGG16\_BN Results}
\label{tab:vgg16}
\end{table*}

\newpage
\section{Conclusion}
\subsection{Significance}
In the present study, a biometric system for cattle identification was developed, utilizing VGG16 and Wide ResNet50 models. Notably, the Wide ResNet50 model achieved a peak accuracy of 99.5\% following image compression. These results are significant for application in a context where high-quality images cannot be obtained owing to high equipment cost as well as field operation costs. In the African context, for example, most farmers are small-holder farmers owning a few tens of cattle as well as herding nomadic communities owning hundreds of cattle. However, it is hard for the farmers or herders to acquire high-quality camera in order to use a biometric identification system, additionally, while agricultural extension officers can obtain a high-definition camera for a larger community, using such cameras require expertise and need careful handling especially when using them in the field. This project shows that even low-quality images can be used to a high degree of accuracy for the biometric identification system therefore requiring no specialized equipment. 

A significant limitation of this approach is scalability. Adding new cattle would require retraining the model in order to learn the new cattle classes, if the number of cattle is many, this could be a significant effort. There are two alternatives to scale the model; 1) scale-up and 2) scale-out. The scale-up approach entails having a single model with all the individual cattle enrolled and as seen, this requires retraining the model in cases of new cattle classes. An alternative strategy proposed in this study involves a scale-out approach, utilizing an ensemble of models to accommodate a larger number of cattle classes. In this configuration, multiple models operate in unison to perform identification tasks. Ensemble models are a common industry practice and that approach can be used to address the scalability issues arising from this methodology.

\subsection{Future Work}
To improve on this work, more models can be considered as implemented in the baseline paper. In addition, an interface can be implemented, such that muzzle images can be uploaded and inference can be made to identify the cattle. This can also be interlaid with a database system that contains more information about the cattle such as the vaccine and disease history, and more. This interface will facilitate a use case such as disease tracking. Some issues related to scalability were also raised during the research. To scale the work, as suggested, an ensemble of models can be used, such that retraining is done on just the newly added cattle classes.

\section*{Acknowledgments}
This work was done in collaboration with CyLab Africa/Upanzi Network and Carnegie Mellon University Africa.

\bibliographystyle{unsrt}  
\bibliography{references}

\end{document}